%% file: iclr2026_conference.tex
\documentclass{article}

\usepackage{microtype}
\usepackage{graphicx}
\usepackage{subfigure}
\usepackage{algorithm}
\usepackage{algorithmic}
\usepackage{booktabs} 

\usepackage{hyperref}
\usepackage{url}
\usepackage{amsmath,amssymb,mathtools}
\usepackage[nameinlink,capitalize,noabbrev]{cleveref}
\crefname{equation}{Eq.}{Eqs.}
\crefname{section}{Section}{Sections}
\crefname{figure}{Figure}{Figures}
\crefname{table}{Table}{Tables}

\newcommand{\DOSS}{\textsc{DOSS}}
\newcommand{\GateDown}{\textsc{GateDown}}
\newcommand{\ObjReturn}{\textsc{Return}}
\newcommand{\ObjDownsideRisk}{\textsc{DownsideRisk}}
\newcommand{\DefaultObj}{\ObjDownsideRisk}
\newcommand{\ObjSharpe}{\textsc{SharpeLike}}

\input{math_commands.tex}

\usepackage{iclr2026_conference,times}

\iclrfinalcopy

\title{Dynamic Objective Selection with Safeguards and LLM Oversight for Financial Decision-Making}

\author{Keigo Sakurai, Takahiro Ogawa, Miki Haseyama \\
Hokkaido University \\
\texttt{sakurai@lmd.ist.hokudai.ac.jp} \\
\texttt{ogawa@lmd.ist.hokudai.ac.jp} \\
\texttt{mhaseyama@lmd.ist.hokudai.ac.jp}
\And
Anjyu Anan \\
Nomura Asset Management Co., Ltd. \\
Kobe University \\
\texttt{an2anju555@gmail.com}
\And
Kei Nakagawa \\
Osaka Metropolitan University \\
\texttt{kei.nak.0315@gmail.com}}

\begin{document}

\maketitle

\begin{abstract}
Financial decision-making tasks such as stock recommendation and portfolio allocation typically estimate future return and risk and then select trades or allocations for an investor, and the chosen optimization objective often determines realized performance.
However, because market conditions evolve over time, a fixed objective can be suboptimal across regimes, while regime-switching pipelines that rely on latent regime estimates can be noisy or delayed and frequent switching can increase turnover and operational instability.
In this paper, we propose DOSS (Dynamic Objective Selection with Safeguards), a learning-based selector that directly chooses the decision-relevant objective function at each time point from interpretable statistical summaries of recent returns, selecting among a small set of candidates (e.g., return-seeking, loss-averse, and risk-adjusted) without introducing intermediate regime variables.
DOSS formulates objective selection as a classification problem over objectives and performs sequential updates with a rolling window to make forward-looking selections without temporal leakage, while also outputting a confidence score for each proposal.
To mitigate misselection and excessive switching in deployment, DOSS applies confidence-aware gating with a fail-safe that overrides low-confidence proposals to a conservative default and enforces explicit controls tied to switching frequency.
We further integrate governance by positioning a Large Language Model (LLM) as an oversight component rather than a generator of new objectives: the LLM is restricted to accept a proposed objective or override it to a predefined safe default, with deterministic rule-based constraints triggering overrides when needed.
\end{abstract}

\section{Introduction}
In financial decision-making tasks such as stock recommendation~\citep{ren2019investment,takayanagi2023personalized,lee2024stock} and portfolio allocation~\citep{pinelis2022machine,imajo2021deep}, a model estimates future return and risk and then selects trades or allocations for an investor.
In such systems, the choice of optimization objective often determines realized performance~\citep{markowitz1952,michaud1989markowitz,lee2011risk}.
However, market conditions evolve over time, and a fixed objective can be suboptimal across regimes.
For example, a return-seeking objective may be suitable in rising markets, whereas a loss-averse or risk-adjusted objective may be preferable in downturns, motivating objective switching over time~\citep{hamilton1989new,nawrocki1999brief,sortino1991downside}.

Despite this need, many AI/ML formulations for finance commit to a single objective (e.g., maximizing return or optimizing a risk criterion) throughout deployment.
Regime-switching pipelines aim to change behavior by inferring latent regimes, but regime estimates are not directly observable and can be noisy or delayed, leading to mistimed switches~\citep{hamilton1989new,dacco1999regime}.
Moreover, frequent switching can increase turnover and operational instability, making objective switching difficult to deploy without explicit controls~\citep{shen2014doubly,nakagawa2021rm}.

A direct formulation is to select the decision-relevant objective function from observable market features at each time point rather than inferring intermediate regime variables.
Such a formulation naturally admits rolling or sequential updates under non-stationary market conditions while preserving a strict forward-looking constraint~\citep{bergmeir2012use,gama2014survey}.
Given interpretable statistical summaries of recent returns, a selector outputs the objective to use from a small set of candidates (e.g., return-seeking, loss-averse, and risk-adjusted), treating objective selection as a policy (a meta-decision) that precedes downstream optimization~\citep{freund1997decision}.

Because objective selection is uncertain, the selector should expose a confidence signal and include a fail-safe that overrides low-confidence proposals to a conservative default, reducing misselection and excessive switching while keeping the basis of selection explicit~\citep{chow2003optimum,guo2017calibration,hendrickx2024machine}.
In deployment, objective switching also requires oversight mechanisms that enforce risk constraints and provide accountable decision traces~\citep{crespo2017evolution}.

To satisfy such deployment requirements, governance can be integrated by positioning a Large Language Model (LLM) as an oversight component rather than a generator of new objectives.
The LLM is restricted to \emph{accept} a proposed objective or \emph{override} it to a predefined safe default, while rule-based constraints (e.g., low confidence or excessive switching) enforce overrides deterministically~\citep{bai2022constitutional,gu2024survey}.
Restricting the LLM to accept/override confines prompt sensitivity and uncertainty to a bounded supervisory role aligned with deployment requirements.

In this paper, we propose \textbf{DOSS} (\underline{D}ynamic \underline{O}bjective \underline{S}election with \underline{S}afeguards), a learning-based selector that chooses among multiple candidate objective functions at each time point from interpretable statistical features summarizing the market state.
DOSS formulates objective selection as a classification problem over objectives and performs sequential updates using a rolling window to make forward-looking selections without temporal leakage.
DOSS also outputs a confidence score and applies a gating mechanism that falls back to a conservative objective under low confidence, mitigating misselection and excessive switching.
Experiments compare DOSS against static objectives, direct LLM-based selection, and a hindsight oracle, showing consistent improvements over the static baseline on a public benchmark.

Our contributions are threefold:
\begin{itemize}
\item We formulate objective-function switching as a policy problem in financial decision-making, where the system selects an objective at each time point under evolving market conditions.
\item We propose DOSS, a learning-based objective selector operating on interpretable statistical features with rolling-window updates for forward-looking selection without temporal leakage.
\item We introduce safeguard and governance mechanisms: confidence-aware gating to a conservative default and optional LLM oversight constrained to accept/override, to control misselection and excessive switching in deployment-oriented settings.
\end{itemize}

\section{Problem Setting}\label{sec:problem_setting}
We study sequential financial decision-making at discrete decision times
$t \in \mathcal{T}=\{1,\dots,T\}$. At each $t$, the system observes an information set $\mathcal{I}_t$
and computes interpretable features $\mathbf{x}_t=\phi(\mathcal{I}_t)\in\mathbb{R}^d$.
Before future outcomes are realized, the system selects an optimization objective $o_t\in\mathcal{O}$
from a small candidate set and then executes a fixed downstream decision module.

\paragraph{Candidate objectives (predefined and bounded).}
Let $\mathcal{O}=\{o^{(1)},\dots,o^{(M)}\}$ be a fixed, predefined candidate set; each $o\in\mathcal{O}$
corresponds to a scalar objective $J_o(\cdot;\mathcal{S}_t)$ used by the downstream optimizer.
Throughout, objective switching selects among this bounded set only (no latent regimes and no expansion);
in experiments we use $\mathcal{O}=\{\ObjReturn,\ObjDownsideRisk,\ObjSharpe\}$ as objectives already supported by the benchmark.

\paragraph{Downstream decision module (fixed).}
At each $t$, the downstream module outputs an action
\begin{equation}
a_t=\textsc{Optimize}(o_t;\mathcal{S}_t)\in\mathcal{A}_t,
\end{equation}
and is held fixed across all compared policies (same solver/seed and constraint/cost modeling), so differences isolate the objective-selection policy.

\paragraph{Realized utility and offline objective sweep.}
Let $U(t,a;H)$ denote the realized utility of executing action $a$ at time $t$ and evaluating over horizon $H$.
For any $o\in\mathcal{O}$, define
\begin{equation}
u(t,o)\triangleq U\!\left(t,\,\textsc{Optimize}(o;\mathcal{S}_t);H\right).
\end{equation}
Because $|\mathcal{O}|$ is small, we can compute $u(t,o)$ for all $o\in\mathcal{O}$ offline (objective sweep).

\paragraph{Hindsight oracle label (for supervised learning only).}
The best-in-hindsight objective within the candidate set is
\begin{equation}
o_t^\star=\arg\max_{o\in\mathcal{O}} u(t,o).
\label{eq:oracle_objective}
\end{equation}

\paragraph{Forward-looking constraint and label availability.}
The policy must be measurable w.r.t.\ $\mathcal{I}_t$: $o_t=\pi(\mathbf{x}_t)$ cannot use future data.
When training online/rolling, the oracle label $o_i^\star$ for a past index $i$ is available at time $t$
only if its evaluation horizon has fully realized (i.e., $i+H\le t$), preventing temporal leakage.

\paragraph{Deployment-oriented safeguards.}
Objective switching can induce instability. We include (i) confidence gating to a conservative default objective $o^{\mathrm{def}}\in\mathcal{O}$
and (ii) optional switching-frequency control. Let $o_t^{\mathrm{raw}}$ be the selector proposal and $c_t\in[0,1]$ a confidence score. The gated objective is
\begin{equation}
o_t=
\begin{cases}
o^{\mathrm{def}}, & c_t<\tau,\\
o_t^{\mathrm{raw}}, & \text{otherwise},
\end{cases}
\label{eq:confidence_gating}
\end{equation}
with threshold $\tau$. We also define a trailing-window switch count over executed objectives:
\begin{equation}
\textsc{SwitchCount}(t;k)=\sum_{j=\max\{2,\,t-k+1\}}^{t} \mathbb{I}[o_j\neq o_{j-1}],
\label{eq:switch_count}
\end{equation}
which can be constrained to limit excessive objective churn in deployment.

\section{Method}\label{sec:method}
\DOSS\ (Dynamic Objective Selection with Safeguards) selects, at each decision time $t$,
a decision-relevant objective $o_t\in\mathcal{O}$ from interpretable features $\mathbf{x}_t$,
without introducing latent regime variables. The method has four components:
(1) oracle label construction for past times (offline only),
(2) a rolling-window objective selector trained as a classifier,
(3) confidence-aware safeguards (gating and optional switching control), and
(4) an optional constrained LLM auditor for governance (accept/override only).
An overview of the proposed procedure is summarized in \cref{alg:doss}.

\subsection{Overview}
At time $t$, \DOSS\ outputs a distribution over candidate objectives $p_\theta(o\mid\mathbf{x}_t)$,
proposes the most likely objective, and then applies deterministic safeguards before passing the final
objective to the downstream optimizer. Rolling-window retraining adapts to evolving market conditions
while preserving a strict forward-looking information constraint.

\subsection{Rolling-window objective selector}
\paragraph{Training data with matured oracle labels.}
Let $H$ be the evaluation horizon used in $U(\cdot,\cdot)$ (cf.\ \cref{sec:problem_setting}).
At decision time $t$, define the set of past indices whose oracle labels have fully realized:
\begin{equation}
\mathcal{I}^{\mathrm{mature}}_t \triangleq \{i:\, t-w \le i \le t-1 \ \ \text{and}\ \ i+H \le t\}.
\end{equation}
We form a rolling training set
$\mathcal{D}_t=\{(\mathbf{x}_i, y_i)\}_{i\in\mathcal{I}^{\mathrm{mature}}_t}$,
where $y_i=o_i^\star$ is the hindsight-best objective in \cref{eq:oracle_objective}.
This ensures that at time $t$ the selector is trained only on supervision available by $t$.
This rolling-window update is a standard approach for adapting to non-stationary environments (concept drift) while keeping the training data recent~\citep{gama2014survey}.

\paragraph{Model and objective.}
We parameterize the selector as a linear softmax classifier:
\begin{equation}
p_\theta(o\mid\mathbf{x})=
\frac{\exp(\mathbf{w}_o^\top\mathbf{x}+b_o)}
{\sum_{o'\in\mathcal{O}}\exp(\mathbf{w}_{o'}^\top\mathbf{x}+b_{o'})},
\end{equation}
and minimize regularized cross-entropy over the rolling window:
\begin{equation}
\mathcal{L}_t(\theta)=
-\frac{1}{|\mathcal{I}^{\mathrm{mature}}_t|}
\sum_{i\in\mathcal{I}^{\mathrm{mature}}_t}\log p_\theta(y_i\mid\mathbf{x}_i)
+\lambda\lVert\theta\rVert_2^2.
\end{equation}
The linear form keeps the selector interpretable and stable under small sample sizes.

\paragraph{Inference.}
Given $\mathbf{x}_t$, the raw proposal is the MAP objective
\begin{equation}
o_t^{\mathrm{raw}}=\arg\max_{o\in\mathcal{O}} p_\theta(o\mid\mathbf{x}_t).
\end{equation}

\subsection{Confidence-aware safeguards}
\paragraph{Confidence score.}
We define confidence as the probability gap between the top-1 and top-2 objectives:
\begin{equation}
c_t = p_{(1)}(\mathbf{x}_t)-p_{(2)}(\mathbf{x}_t),
\end{equation}
where $p_{(1)}(\mathbf{x}_t)$ and $p_{(2)}(\mathbf{x}_t)$ are the largest and second-largest
entries of $p_\theta(\cdot\mid\mathbf{x}_t)$.

\paragraph{Gating (fail-safe default).}
We apply confidence gating to a conservative default objective $o^{\mathrm{def}}$ via
\cref{eq:confidence_gating}. This explicitly mitigates misselection when the selector is uncertain,
and reduces instability by avoiding low-confidence switches into aggressive objectives.
This is closely related to decision-making with a reject option, where uncertain predictions are abstained or mapped to a safe fallback~\citep{chow2003optimum,hendrickx2024machine}.

\paragraph{Optional switching-frequency control.}
To further limit operational churn, we optionally enforce a threshold on
$\textsc{SwitchCount}(t;k)$ in \cref{eq:switch_count}. If the threshold is exceeded, the objective is
overridden to $o^{\mathrm{def}}$. This safeguard is deterministic and auditable.

\subsection{Optional constrained LLM auditor (governance)}
In deployment, \DOSS\ can be augmented with an LLM used strictly as an auditor, not as an objective generator.
The auditor observes $(\mathbf{x}_t, \tilde{o}_t, c_t)$ \emph{after} deterministic safeguards and outputs \textsc{Accept} or \textsc{Override} plus a short rationale.
If \textsc{Override}, the final objective is set to $o^{\mathrm{def}}$.
This keeps the decision space bounded ($\mathcal{O}$) and yields an auditable accept/override trace; we do not evaluate the auditor in the candidate-only experiments.

\begin{algorithm}[t]
\caption{\DOSS\ with safeguards (and optional LLM auditor)}\label{alg:doss}
\begin{algorithmic}[1]
\STATE \textbf{Inputs:} candidate objectives $\mathcal{O}$, default $o^{\mathrm{def}}$, window $w$,
horizon $H$, reg.\ $\lambda$, confidence threshold $\tau$, switch window $k$ and max switch $s_{\max}$.
\STATE \textbf{For} $t=1$ to $T$:
\STATE \quad Observe info set $\mathcal{I}_t$ and features $\mathbf{x}_t=\phi(\mathcal{I}_t)$; form state $\mathcal{S}_t$.
\STATE \quad Construct matured index set $\mathcal{I}^{\mathrm{mature}}_t=\{i:\ t-w \le i \le t-1,\ i+H\le t\}$.
\STATE \quad \textbf{(Train)} Fit $\theta_t=\arg\min_\theta \mathcal{L}_t(\theta)$ on $\{(\mathbf{x}_i,o_i^\star)\}_{i\in\mathcal{I}^{\mathrm{mature}}_t}$.
\STATE \quad \textbf{(Predict)} Compute $p_{\theta_t}(\cdot\mid\mathbf{x}_t)$ and $o_t^{\mathrm{raw}}=\arg\max_{o\in\mathcal{O}} p_{\theta_t}(o\mid\mathbf{x}_t)$.
\STATE \quad \textbf{(Confidence)} Let $c_t=p_{(1)}(\mathbf{x}_t)-p_{(2)}(\mathbf{x}_t)$.
\STATE \quad \textbf{(Gating)} Set $\tilde{o}_t=o^{\mathrm{def}}$ if $c_t<\tau$, else $\tilde{o}_t=o_t^{\mathrm{raw}}$.
\STATE \quad \textbf{(Switch control, optional)} If $\textsc{SwitchCount}(t;k)>s_{\max}$ then set $\tilde{o}_t=o^{\mathrm{def}}$.
\STATE \quad \textbf{(LLM audit, optional)} Query auditor with $(\mathbf{x}_t,\tilde{o}_t,c_t)$; if \textsc{Override} then set $o_t=o^{\mathrm{def}}$ else $o_t=\tilde{o}_t$.
\STATE \quad \textbf{(Execute)} Output action $a_t=\textsc{Optimize}(o_t;\mathcal{S}_t)$.
\STATE \textbf{End for}
\end{algorithmic}
\end{algorithm}

\section{Experiments}\label{sec:experiments}
\subsection{Experimental Settings}\label{sec:setting}
\subsubsection{Benchmark and evaluation protocol}\label{sec:exp_protocol}
We instantiate \DOSS\ on FAR-Trans~\citep{sanz2024far}, a public benchmark for financial asset recommendation with rolling evaluation over market time.
Following the benchmark protocol, we evaluate over $|\mathcal{T}|=61$ decision time points.
Each time point defines a split where (i) market and transaction information prior to $t$ forms the available information,
and (ii) outcomes in the subsequent $(t,t+\Delta t)$ interval form the evaluation period, with $\Delta t=6$ months and time points spaced two weeks apart.%
\footnote{We use FAR-Trans' standard rolling construction: 61 dataset variants, $\Delta t=6$ months, two-week spacing.}

In this instantiation, the evaluation horizon in Section~\ref{sec:problem_setting} corresponds to a six-month interval.
We denote by $H$ the number of decision steps spanning this interval (i.e., $H$ is in decision-time units consistent with the index arithmetic in \cref{sec:method}).
Therefore, at time $t$, oracle labels are available only for past indices whose six-month evaluation windows have fully realized, matching the matured-label constraint in Section~\ref{sec:method}.

We treat each benchmark time point as a decision time $t$ in \cref{sec:problem_setting}.

\subsubsection{Candidate objectives and fixed downstream module}\label{sec:exp_candidate_space}
We evaluate in the \textbf{candidate-only} setting: the downstream decision module $\textsc{Optimize}(\cdot)$ and its constraints are fixed, and only the objective-selection policy changes.
We instantiate the bounded candidate set as $\mathcal{O}=\{\ObjReturn,\ObjDownsideRisk,\ObjSharpe\}$, representing return-seeking, downside-risk (loss-averse), and risk-adjusted (Sharpe-like), respectively.
At each $t$, the downstream module receives $(o_t,\mathcal{S}_t)$ and outputs a feasible action (e.g., portfolio weights or a ranked list) consistent with implementation constraints (budget/long-only/position limits, and any transaction-cost terms if included).
For offline analysis, we compute $u(t,o)$ for all $o\in\mathcal{O}$ by running the fixed downstream module under each candidate objective.

\subsubsection{Objective-selection metric and stability statistics}\label{sec:exp_metrics}
Let $u(t,o)$ be the realized utility over the benchmark evaluation horizon when objective $o$ is used at $t$
(see \cref{sec:problem_setting}).

We measure objective-selection quality via the hit indicator with tolerance $\epsilon>0$.
Let $o_t=\pi(\mathbf{x}_t)$ be the objective selected by policy $\pi$ at time $t$.
Then
\begin{equation}
\mathrm{hit}(t;\pi)=\mathbb{I}\!\left[u(t,o_t) \ge \max_{o\in\mathcal{O}}u(t,o)-\epsilon\right].
\end{equation}

and report Hit Percentage Accuracy (HPA):
\begin{equation}
\textsc{HPA}(\pi)=\frac{1}{|\mathcal{T}|}\sum_{t\in\mathcal{T}}\mathrm{hit}(t;\pi).
\end{equation}
We use $\epsilon=0.02$ in all experiments.

Because deployment concerns include operational instability, we also report:
(i) \textbf{switch rate} $=\frac{1}{|\mathcal{T}|-1}\sum_{t>1}\mathbb{I}[o_t\neq o_{t-1}]$ and
(ii) \textbf{gating rate} $=\frac{1}{|\mathcal{T}|}\sum_{t}\mathbb{I}[c_t<\tau]$ for confidence gating (applicable to gated policies such as \DOSS--\GateDown).

\subsubsection{Compared policies}\label{sec:exp_policies}
We compare:
\begin{itemize}
\item \textbf{\DOSS--\GateDown}: \DOSS\ with confidence gating to a conservative default $o^{\mathrm{def}}=\DefaultObj$ and threshold $\tau$.
Unless stated otherwise, we report a representative operating point $\tau=0.2$ (and include a threshold sweep).
\item \textbf{Static}($o$): fixed objective $o\in\mathcal{O}$ for all $t$.
\item \textbf{LLM-Direct (GPT-4.1-mini)}: an LLM directly predicts an objective in $\mathcal{O}$ from the same interpretable features $\mathbf{x}_t$.
We use \texttt{gpt-4.1-mini} via the OpenAI API (accessed on \texttt{2026-02-05}), with temperature $0.0$ (and top\_p $=1.0$) for determinism.
The prompt provides $\mathbf{x}_t$ and instructs the model to output \emph{exactly one} label in
\{\texttt{return}, \texttt{downside\_risk}, \texttt{sharpe\_like}\}.
\item \textbf{LLM-Direct (Mock)}: a deterministic heuristic mapping from $\mathbf{x}_t$ to an objective in $\mathcal{O}$,
used as a non-stochastic proxy baseline.
\item \textbf{Oracle}: the hindsight upper bound $o_t^\star=\arg\max_{o\in\mathcal{O}}u(t,o)$ (offline only).
\end{itemize}
Our main results are for objective selection; the optional constrained LLM auditor in \cref{sec:method} is a deployment extension and is not required for the candidate-only comparisons.

\subsubsection{Implementation details}\label{sec:exp_impl}
We use a rolling window size $w=30$ time points.
At each decision time $t$, \DOSS\ retrains the linear softmax selector on matured labels (only indices $i$ whose evaluation horizon has realized), as specified in \cref{sec:method}.
We set $\lambda=0.01$ for $\ell_2$ regularization.
For confidence gating, we sweep $\tau\in\{0.0,0.1,0.2,0.3,0.4\}$ and report both HPA and gating rate.
\paragraph{LLM prompting.}
For LLM-Direct (GPT-4.1-mini), we format $\mathbf{x}_t$ as a compact JSON dictionary of feature names/values and request exactly one label in the bounded set
\{\texttt{return}, \texttt{downside\_risk}, \texttt{sharpe\_like}\} with temperature $0.0$.
We map these labels to objectives in $\mathcal{O}$ as
\texttt{return}$\mapsto \ObjReturn$, \texttt{downside\_risk}$\mapsto \ObjDownsideRisk$, and \texttt{sharpe\_like}$\mapsto \ObjSharpe$.

\subsection{Results}\label{sec:results}
\subsubsection{Main results (candidate-only)}\label{sec:exp_main}
Table~\ref{tab:main_candidate} reports both objective-selection accuracy (HPA) and operational stability measured by switch rate.
Our main configuration, \DOSS--\GateDown\ ($\tau=0.2$), achieves $\textsc{HPA}=0.5626$, improving over the strongest static objective Static(\ObjDownsideRisk) ($0.5235$) by $+0.0391$ absolute (about $+7.5\%$ relative).
Because the candidate set $\mathcal{O}$ and the downstream decision module are fixed, this difference is attributable to the objective-selection policy.

\begin{table}[t]
\centering
\caption{Candidate-only objective-selection results on FAR-Trans ($|\mathcal{T}|=61$, $\epsilon=0.02$).
HPA measures how often the selected objective is within $\epsilon$ of the best candidate objective at each time point.
Switch rate is the fraction of adjacent time points where the selected objective changes.
Figure~\ref{fig:hpa_timeline} plots the cumulative HPA counterpart of Table~\ref{tab:main_candidate}.
}
\label{tab:main_candidate}
\begin{tabular}{lcc}
\toprule
Policy & HPA & Switch rate \\
\midrule
\DOSS--\GateDown\ ($\tau=0.2$) & 0.5626 & 0.4590 \\
\textbf{Static}($\ObjDownsideRisk$) & 0.5235 & 0.0000 \\
\textbf{Static}($\ObjReturn$) & 0.4765 & 0.0000 \\
\textbf{Static}($\ObjSharpe$) & 0.4850 & 0.0000 \\
\textbf{LLM-Direct}(\textsc{GPT-4.1-mini}) & 0.5173 & 0.5902 \\
\textbf{LLM-Direct}(\textsc{Mock}) & 0.5433 & 0.4754 \\
\textbf{Oracle} & 0.6150 & 0.8197 \\
\bottomrule
\end{tabular}
\end{table}
A key deployment consideration is whether improved accuracy comes at the cost of excessive switching.
The hindsight oracle attains the highest HPA ($0.6150$) but exhibits a high switch rate ($0.8197$), reflecting aggressive hindsight switching that is typically infeasible or undesirable in practice.
In contrast, \DOSS--\GateDown\ achieves a substantially lower switch rate ($0.4590$) while closing a meaningful fraction of the oracle gap.

\begin{figure}[t]
\centering
\includegraphics[width=0.9\linewidth]{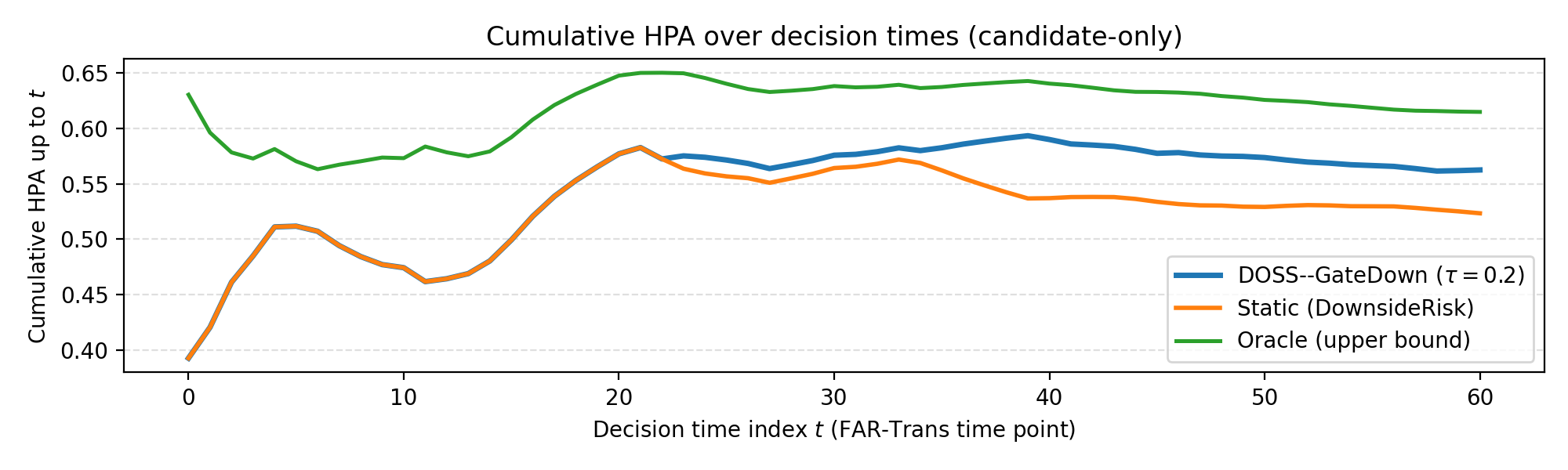}
\caption{Cumulative objective-selection accuracy over time (candidate-only).
We report the cumulative HPA up to each time index for \DOSS--\GateDown\ ($\tau=0.2$),
the strongest static baseline Static(\ObjDownsideRisk), and the hindsight oracle (upper bound).}
\label{fig:hpa_timeline}
\end{figure}

Compared with the direct LLM selector, \DOSS\ achieves higher HPA ($0.5626$ vs.\ $0.5173$) with lower switching ($0.4590$ vs.\ $0.5902$).
Figure~\ref{fig:hpa_timeline} provides a complementary view by plotting the cumulative HPA over time.

\subsubsection{Sensitivity to confidence threshold}\label{sec:exp_tau}
Table~\ref{tab:tau_sweep} characterizes the accuracy--conservatism trade-off governed by the confidence threshold $\tau$.
Disabling gating ($\tau=0$) reduces HPA to $0.5191$, indicating that acting on all raw proposals can be harmful when the selector is uncertain.
A moderate threshold ($\tau=0.2$) yields the best HPA ($0.5626$) while gating at a rate of $0.41$, supporting the interpretation of gating as a fail-safe that selectively rejects low-confidence switches.

\begin{table}[t]
\centering
\caption{Sensitivity of \DOSS\ to the confidence threshold $\tau$ under confidence gating (\GateDown).
Gating rate is the fraction of time points where the selector falls back to the conservative default $o^{\mathrm{def}}=\DefaultObj$.
All entries use the same downstream module and candidate set; only the objective-selection policy changes.
}
\label{tab:tau_sweep}
\begin{tabular}{ccc}
\toprule
$\tau$ & HPA & Gating rate \\
\midrule
0.0 & 0.5191 & 0.00 \\
0.1 & 0.5491 & 0.24 \\
0.2 & 0.5626 & 0.41 \\
0.3 & 0.5439 & 0.56 \\
0.4 & 0.5155 & 0.74 \\
\bottomrule
\end{tabular}
\end{table}

When $\tau$ becomes too large, gating dominates the decision rule: the selector falls back to the conservative default too often (gating rate $0.74$ at $\tau=0.4$), effectively collapsing toward a near-static strategy and losing adaptivity (HPA $0.5155$).
Overall, the sweep exhibits a natural ``U-shaped'' behavior: insufficient gating admits noisy switches, whereas overly conservative gating removes the benefit of objective switching.

\subsubsection{Qualitative behavior over time}\label{sec:exp_qual}
\begin{figure}[t]
\centering
\includegraphics[width=0.9\linewidth]{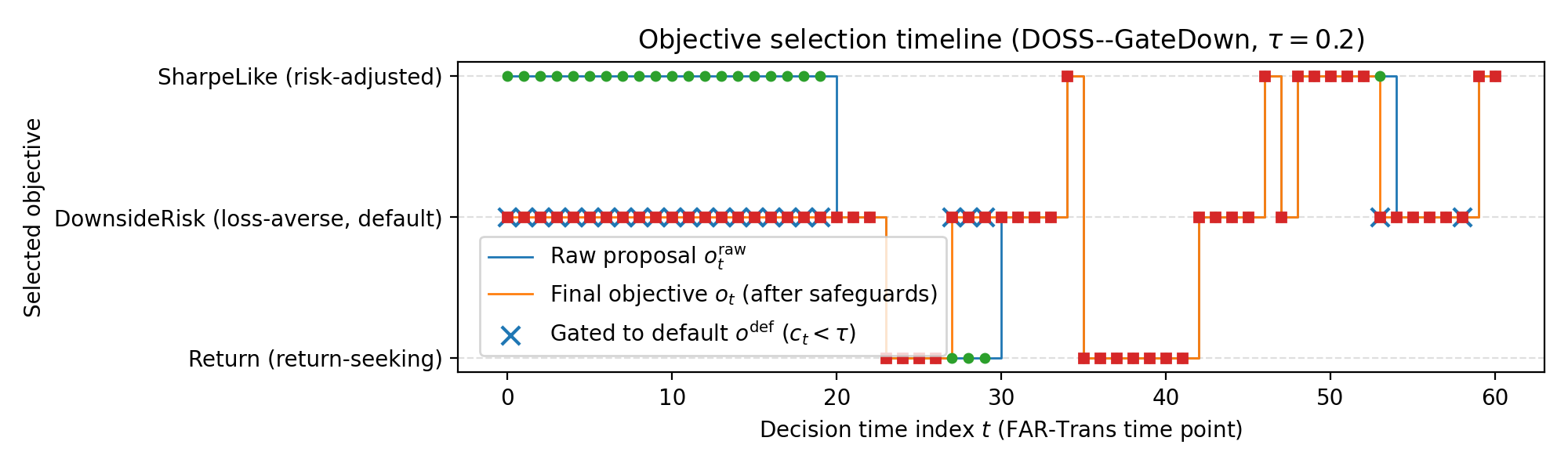}
\caption{Objective-selection timeline for \DOSS--\GateDown\ ($\tau=0.2$).
We plot the raw proposal $o_t^{\mathrm{raw}}$ and the final objective $o_t$ after deterministic safeguards.
Markers indicating ``override'' correspond to time points where safeguards replace the raw proposal with the conservative default $o^{\mathrm{def}}$
(e.g., triggered by confidence gating when $c_t<\tau$).}
\label{fig:objective_timeline}
\end{figure}

Figure~\ref{fig:objective_timeline} visualizes both the raw selector proposals $o_t^{\mathrm{raw}}$ and the final objectives $o_t$ after safeguards for \DOSS--\GateDown\ ($\tau=0.2$).
The two series frequently coincide, indicating that when the selector is confident, its proposed objective is typically executed as-is.
Override markers highlight time points where safeguards replace the raw proposal with the conservative default $o^{\mathrm{def}}$, preventing potentially unstable transitions under uncertainty.
Together with the moderate switch rate in Table~\ref{tab:main_candidate}, this suggests that \DOSS\ does not ``chase'' short-lived fluctuations but instead switches objectives selectively.

Figure~\ref{fig:hpa_timeline} shows that \DOSS\ maintains an advantage over the strongest static baseline across the evaluation window, rather than relying on a few isolated time points.

\subsubsection{Statistical reliability: bootstrap over time points}\label{sec:exp_bootstrap}
We assess whether the improvement over the strongest static baseline is robust across time points.
Let $\Delta_t=\mathrm{hit}(t;\pi_{\mathrm{DOSS}})-\mathrm{hit}(t;\pi_{\mathrm{StaticDown}})$ and $\bar{\Delta}=\frac{1}{|\mathcal{T}|}\sum_t\Delta_t$.
We perform a nonparametric bootstrap by sampling time points with replacement and recomputing $\bar{\Delta}$.
Table~\ref{tab:bootstrap} reports a mean improvement of $\bar{\Delta}=0.0391$ in HPA, a 95\% bootstrap confidence interval of $[0.0118,\,0.0706]$, and $P(\bar{\Delta}>0)=0.9986$.

We note that time points are not strictly i.i.d.\ due to temporal dependence; we interpret the bootstrap as a pragmatic robustness check on whether the improvement is driven by a small number of outliers.

\begin{table}[t]
\centering
\caption{Bootstrap over time points for \DOSS--\GateDown\ ($\tau=0.2$) vs.\ the strongest static baseline (candidate-only).
We report mean $\Delta$HPA, the 95\% bootstrap confidence interval, and the probability that the \emph{mean} improvement is positive.
$P(\bar{\Delta}>0)$ is estimated from the bootstrap replicates of the mean difference.
}
\label{tab:bootstrap}
\scalebox{0.88}{
\begin{tabular}{lccc}
\hline
Comparison & Mean $\Delta$HPA & 95\% CI & $P(\bar{\Delta}>0)$ \\
\hline
\DOSS--\GateDown\ ($\tau=0.2$) vs \textbf{Static}($\ObjDownsideRisk$)
& 0.0391 & [0.0118, 0.0706] & 0.9986 \\
\hline
\end{tabular}
}
\normalsize
\end{table}

\subsubsection{Discussion: bounded adaptivity under operational constraints}\label{sec:exp_discuss}
The results illustrate a practical tension in financial deployment: adaptivity can improve outcomes, but frequent behavioral changes can be costly.
Static policies incur zero switching but cannot react to evolving market conditions, whereas the oracle achieves the highest HPA with a very high switch rate.
\DOSS\ provides a controlled middle ground: it captures a meaningful portion of the oracle headroom while keeping switching substantially lower, and the confidence gating sweep shows that a moderate level of conservative fallback improves accuracy rather than merely reducing churn.

A further benefit of objective switching in a bounded candidate set is traceability.
Each decision time produces an auditable record consisting of (i) interpretable feature summaries $\mathbf{x}_t$,
(ii) the selected objective label $o_t\in\mathcal{O}$, and (iii) the confidence signal used to trigger fallbacks.
This structure aligns with governance requirements and motivates the optional LLM auditor in Section~\ref{sec:method} as a supervised accept/override layer that does not expand the decision space beyond $\mathcal{O}$.

\section{Conclusion}\label{sec:conclusion}
We studied objective-function switching as a policy problem in financial decision-making:
at each decision time, select a decision-relevant objective from a small predefined candidate set using only observable,
interpretable summaries of recent returns.
We proposed \DOSS, a rolling-window objective selector with confidence-aware safeguards that (i) expose uncertainty via a confidence score and (ii) provide deployment-oriented fail-safes through deterministic gating (and optional switching control).
In a candidate-only evaluation on FAR-Trans, \DOSS\ improves objective-selection accuracy over static objectives and a direct LLM selector, while explicitly controlling operational instability through conservative fallback.
A constrained LLM auditor can be layered for governance without expanding the action space beyond $\mathcal{O}$.

\bibliography{iclr2026_conference}
\bibliographystyle{iclr2026_conference}

\appendix

\end{document}

%% file: math_commands.tex
\usepackage{amsmath,amsfonts,bm}

\def\eqref#1{equation~\ref{#1}}

\def\1{\bm{1}}

\DeclareMathAlphabet{\mathsfit}{\encodingdefault}{\sfdefault}{m}{sl}
\SetMathAlphabet{\mathsfit}{bold}{\encodingdefault}{\sfdefault}{bx}{n}

